\documentclass{article}
\usepackage[preprint]{neurips_2019}
   \usepackage[pdftex]{graphicx}     
   \graphicspath{{../figures/}{./figures/}{./image/}}    
   \DeclareGraphicsExtensions{.pdf,.jpeg,.png,.jpg}   

\usepackage[utf8]{inputenc} 
\usepackage[T1]{fontenc}    
\usepackage{hyperref}       
\usepackage{url}            
\usepackage{booktabs}       
\usepackage{amsfonts}       
\usepackage{nicefrac}       
\usepackage{microtype}      

\title{Human-like machine thinking: Language guided imagination}

\author{%
  Feng Qi\\
  Nuffield Department of Clinical Neurosciences\\
  University of Oxford,UK\\
  \texttt{feng.qi@ndcn.ox.ac.uk} \\
  \And
  Wenchuan Wu \\
  Nuffield Department of Clinical Neurosciences\\
  University of Oxford,UK\\
   \texttt{wenchuan.wu@ndcn.ox.ac.uk} \\
}

\begin{document}

\maketitle

\begin{abstract}
Human thinking requires the brain to understand the meaning of language and to properly organize the thoughts flow using the language. However, current natural language processing models are primarily limited to the word probability estimation. Here, we proposed a Language Guided Imagination (LGI) network to incrementally learn the meaning and usage of diverse words and syntaxes, aiming to form a human-like machine thinking process. LGI contains three subsystems: (1) vision system that contains an encoder to disentangle the input or imagined scenarios into abstract population representations, and an imagination decoder to reconstruct imagined scenarios from higher level representations; (2) Language system, which consists of a binarizer to transfer symbol texts into binary vectors, an IPS (mimicking the human IntraParietal Sulcus, implemented by an LSTM) to extract the quantity information from the input texts, and a textizer to convert binary vectors into text symbols; (3) a PFC (mimicking the human PreFrontal Cortex, implemented by an LSTM) that combines inputs in the forms of both language and vision, and predict text symbols and manipulated images accordingly. In this work, the proposed LGI network illustrates the ability to incrementally learn eight different syntaxes and form a machine thinking loop that enables interactions between language and vision system, which hasn't been demonstrated before. The paper presents a new architecture that allows the machine to learn, understand and use language in a human-like way, which might ultimately enable a machine to construct fictitious 'mental' scenario and possess intelligence.
\end{abstract}

\section{Introduction}
Human thinking is regarded as ‘mental ideas flow guided by language to achieve a goal’. For instance, after seeing heavy rain, you may say internally ‘holding an umbrella could avoid getting wet’, and then you will take an umbrella before leaving. In the process, we know that the visual input of ‘water drop’ is called rain, and can imagine ‘holding an umbrella’ could keep off the rain, and can even experience the feeling of being wet. This continual thinking capacity distinguishes us from the machine, even though the latter can also recognize images, process language, and sense rain-drops. Continual thinking requires the capacity to generate mental imagination guided by language, and extract language representations from a real or imagined scenario.

Modern natural language processing (NLP) techniques can handle question answering etc. tasks, such as answering that ‘Cao Cao’s nickname is Meng De’ based on the website knowledge [1]. However, the NLP network is just a probability model [2] and does not know whether Cao Cao is a man or cat. Indeed, it even does not understand what is a man. On the other hand, human being learns Cao Cao with his nickname via watching TV. When presented the question ‘what’s Cao Cao’s nickname?’, we can give the correct answer of ‘Meng De’ while imagining the figure of an actor in the brain. In this way, we say the machine network does not understand it, but the human does.

Human beings possess such thinking capacity due to its cumulative learning capacity accompanying the neural developmental process. Initially, parent points to a real apple and teaches the baby ‘this is an apple’. After gradually assimilating the basic meanings of numerous nouns, children begin to learn some phrases and finally complicated syntaxes. Unlike the cumulative learning, most NLP techniques normally choose to learn by reading and predicting target words. After consuming billions of words in corpus materials [2], the NLP network can predict ‘Trump’ following ‘Donald’, but it is merely a probability machine.

The human-like thinking system often requires specific neural substrates to support the corresponding functionalities. The most important brain area related to thinking is the prefrontal cortex (PFC), where the working memory takes place, including but not confined to, the maintenance and manipulation of particular information [3]. With the PFC, human beings can analyze and execute various tasks via ‘phonological loop’ and ‘visuospatial scratchpad’ etc. [4,5]. Inspired by the human-like brain organization, we build a ‘PFC’ network to combine language and vision streams to achieve tasks such as language controlled imagination, and imagination based thinking process. Our results show that the LGI network could incrementally learn eight syntaxes rapidly. Based on the LGI, we present the first language guided continual thinking process, which shows considerable promise for the human-like strong machine intelligence.

\section{Related work}
Our goal is to build a human-like neural network by removing components unsupported by neuroscience from AI architecture while introducing novel neural mechanisms and algorithms into it. Taking the convolution neural network (CNN) as an example, although it has reached human-level performance in image recognition tasks [6], animal neural systems do not support such kernel scanning operation across retinal neurons, and thus the neuronal responses measured on monkeys do not match that of CNN units [7,8]. Therefore, instead of CNN, we used fully connected (FC) module [9] to build our neural network, which achieved more resemblance to animal neurophysiology in term of the network development, neuronal firing patterns, object recognition mechanism, learning and forgetting mechanisms, as illustrated in our concurrent submission [10]. In addition, the error backpropagation technique is generally used to modify network weights to learn representation and achieve training objectives [11]. However, in neuroscience, it is the activity-dependent molecular events (e.g. the inflow of calcium ion and the switching of glutamate N-methyl-D-aspartate receptor etc.) that modify synaptic connections [12, 13]. Indeed, the real neural feedback connection provides the top-down imagery information [14], which is usually ignored by AI network constructions due to the concept of error backpropagation. What’s more, our concurrent paper [10] demonstrates that the invariance property of visual recognition under the rotation, scaling, and translation of an object is supported by coordinated population coding rather than the max-pooling mechanism [15]. The softmax classification is usually used to compute the probability of each category (or word) in the repository (or vocabulary) before prediction. However, in reality, we never evaluate all fruit categories in mind before saying ‘it is an apple’, let alone the complicated computation of the normalization term in the softmax. In this paper, we demonstrate object classification is directly output by neurons via a simple rounding operation, rather than the neuroscience unsupported softmax classification [16].

Modern autoencoder techniques could synthesize an unseen view for the desired viewpoint. Using car as an example [17], during training, the autoencoder learns the 3D characteristics of a car with a pair of images from two views of the same car together with the viewpoint of the output view. During testing, the autoencoder could predict the desired image from a single image of the car given the expected viewpoint. However, this architecture is task-specific, namely that the network can only make predictions on cars' unseen views. To include multiple tasks, we added an additional PFC layer that can receive task commands conveyed via language stream and object representation via the visual encoder pathway, and output the modulated images according to task commands and the desired text prediction associated with the images. In addition, by transmitting the output image from the decoder to the encoder, an imagination loop is formed, which enables the continual operation of a human-like thinking process involving both language and image.

\section{Architecture}

\begin{figure}[!htbp]
  \centering\includegraphics[width=1\textwidth]{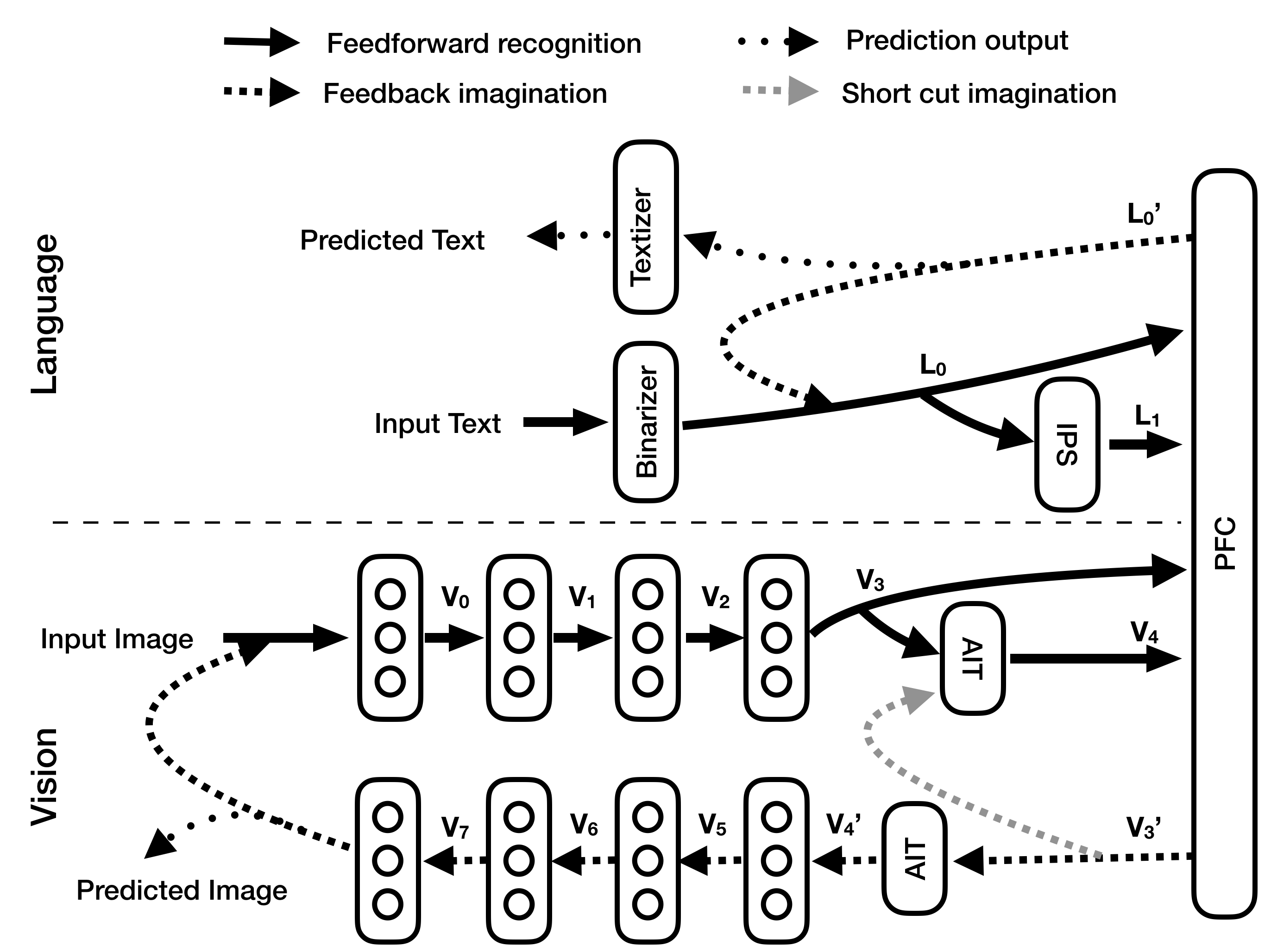}
  \caption[test1.]{The LGI architecture. It contains three subsystems that are trained separated. In the vision subsystem, the encoder can transfer an input or imagined (or predicted) image into a population representation vector \({\bf V_4}\) at the AIT layer (mimicking the Anterior Temporal Lobe for high-level image representation), and the decoder can reconstruct a \({\bf  V_3'}\) output from PFC to a predicted image, which can be fed into the encoder to form the imagination loop. In the language subsystem, a binarizer can transfer the input text symbols into binary representation vectors \({\bf  L_0}\), and a texitizer can transfer the predicted vector \({\bf  L_0'}\) from the PFC into predicted text symbols, which can also be fed into the language loop. There is an IPS layer implemented by an LSTM to extract quantity information \({\bf  L_1}\) from the text vector \({\bf L_0}\). The PFC layer serves as working memory, that takes the concatenated input \({\bf  [L_0, L_1, V_3, V_4]}\) from both language and vision subsystems, and output the predicted next frame representation that could be fed back into both subsystems to form an imagination loop. LIG can use the short cut imagination path (rendered in grey) to rapidly feel the predicted scenario without fully reconstructing the predicted images. }
\end{figure}

As is shown in Figure 1, the LGI network contains three main subsystems including the vision, language and PFC subsystems. The vision autoencoder network was trained separately, whose characteristics of development, recognition, learning, and forgetting can be referred to [10]. After training, the autoencoder is separated into two parts: the encoder (or recognition) part ranges from the image entry point to the final encoding layer, which functions as human anterior inferior temporal lobe (AIT) to provide the high-level abstract representation of the input image [18]; the decoder (or imagination) part ranges from the AIT to image prediction point. The activity vectors of the third encoding layer \({\bf V_3}\) and AIT layer \({\bf V_4}\) are concatenated with language activity vectors \({\bf [L_0, L_1]}\) as input signals to the PFC. We expect, after acquiring the language command, the PFC could output a desired visual activation vector \({\bf V_3’}\), based on which the imagination network could reconstruct the predicted image. Finally, the predicted or imagined image is fed back to the encoder network for the next thinking iteration.

The language processing component first binarizes the input text symbol-wise into a sequence of binary vectors \({\bf [L_0}(t=0), …, {\bf L_0}(T)]\), where T is the text length. To improve the language command recognition, we added one LSTM layer to extract the quantity information of the text (for example, suppose text = ‘move left 12’, the expected output \({\bf L_1}\) is 1 dimensional quantity 12 at the last time point). This layer mimics the number processing functionality of human Intra-Parietal Sulcus (IPS), so it is given the name IPS layer. The PFC outputs the desired activation of \({\bf L_0’}(t)\), which can either be decoded by the ‘texitizer’ into predicted text or serve as \({\bf L_0}(t+1)\) for the next iteration of the imagination process. Here, we propose a textizer (a rounding operation, followed by symbol mapping from binary vector, whose detailed discussion can be referred to the Supplementary section A) to classify the predicted symbol instead of softmax operation which has no neuroscience foundation.

\begin{figure}[!htbp]
  \centering\includegraphics[width=1\textwidth]{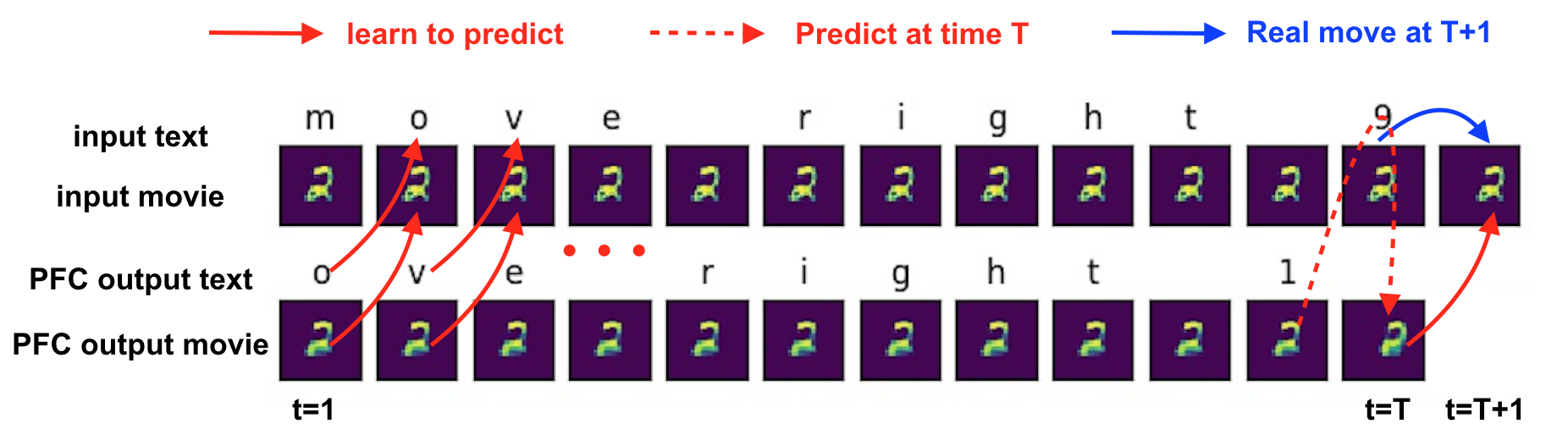}
  \caption[test1.]{Training based on the next frame prediction (NFP). The LSTM-like PFC is trained by the NFP principle, where the goal of the PFC is to output the representation vectors (including both language and vision) of the next frame, as indicated by red arrows. The red dash arrow indicates that, at time T, the PFC of LGI curately generated the mentally manipulated digit instance, which required the understanding of the previous text language and observed images.
}
\end{figure}

The PFC subsystem contains a LSTM and a full connected layer. It receives inputs from both language and vision subsystems in a concatenated form of \({\bf c}(t)=[{\bf L_0}(t), {\bf L_1}(t), {\bf V_3}(t), {\bf V_4}(t)]\) at time t, and gives a prediction output \({\bf a’}(t)=[{\bf L_0’}(t), {\bf V_3’}(t)]\), which is expected to be identical to \({\bf a}(t+1)= [{\bf L_0(}t+1), {\bf V_3}(t+1)]\) at time t+1. This has been achieved with a next frame prediction (NFP) loss function as, \(Loss = \sum_{t=1}^{T-1} ||{\bf a'}(t)-{\bf a}(t+1)||^2/(T-1)\). So given an input image, the PFC can predict the corresponding text description; while given an input text command the PFC can predict the corresponding manipulated image. This NFP loss function has neuroscience foundation, since the molecular mediated synaptic plasticity always takes place after the completion of an event, when the information of both t and t+1 time points have been acquired and presented by the neural system. The strategy of learning by predicting its own next frame is essentially an unsupervised learning.

For human brain development, the visual and auditory systems mature in much earlier stages than the PFC [19]. To mimic this process, our PFC subsystem was trained separately after vision and language components had completed their functionalities. We have trained the network to accumulatively learn eight syntaxes, and the related results are shown in the following section.  Finally, we demonstrate how the network forms a thinking loop with text language and imagined pictures.

\section{Experiment}
The first syntaxes that LGI has learned are the ‘move left’ and ‘move right’ random pixels, with the corresponding results shown in Figure 3. After 50000 steps training, LGI could not only reconstruct the input image with high precision but also predict the 'mentally' moved object with specified morphology, correct manipulated direction and position just after the command sentence completed. The predicted text can complete the word ‘move’ given the first letter ‘m’ (till now, LGI has only learned syntaxes of ‘move left or right’). LGI tried to predict the second word ‘right’ with initial letter ‘r’, however, after knowing the command text is ‘l’, it turned to complete the following symbols with ‘eft’. It doesn’t care if the sentence length is 12 or 11, the predicted image and text just came at proper time and position. Even if the command asked to move out of screen, LGI still could reconstruct the partially occluded image with high fidelity.

\begin{figure}[!htbp]
  \centering\includegraphics[width=1\textwidth]{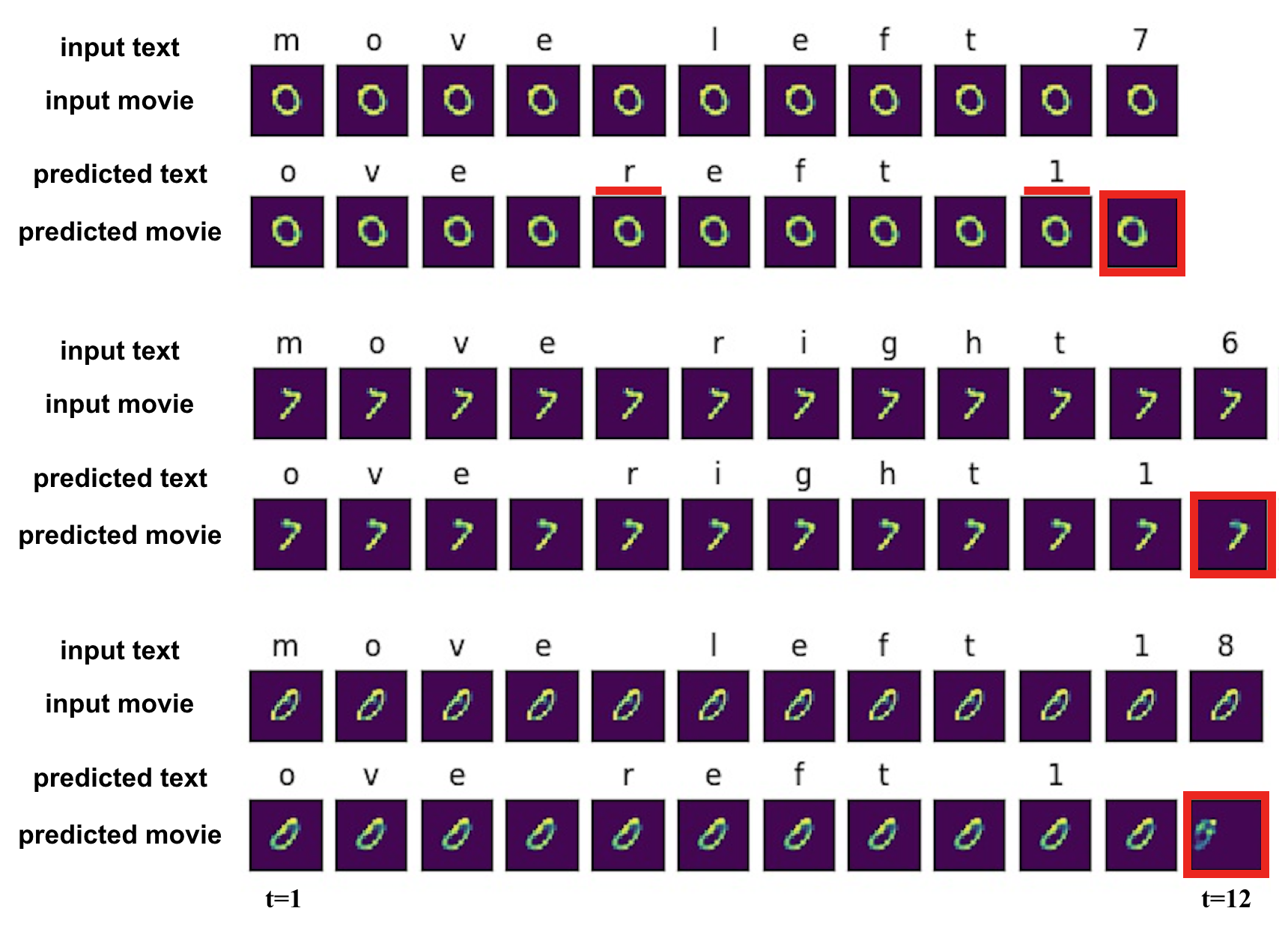}
  \caption[test1.]{Mental manipulation of images based on syntaxes of ‘move left x’ and ‘move right x’, where x is a random number, ranging from 0 to 28. LGI has the capacity to correctly predict the next text symbols and image manipulation (with correct morphology, position, direction) at the proper time point. It can recognize the sentence with flexible text length and digit length.
}
\end{figure}

\begin{figure}[!htbp]
  \centering\includegraphics[width=0.75\textwidth]{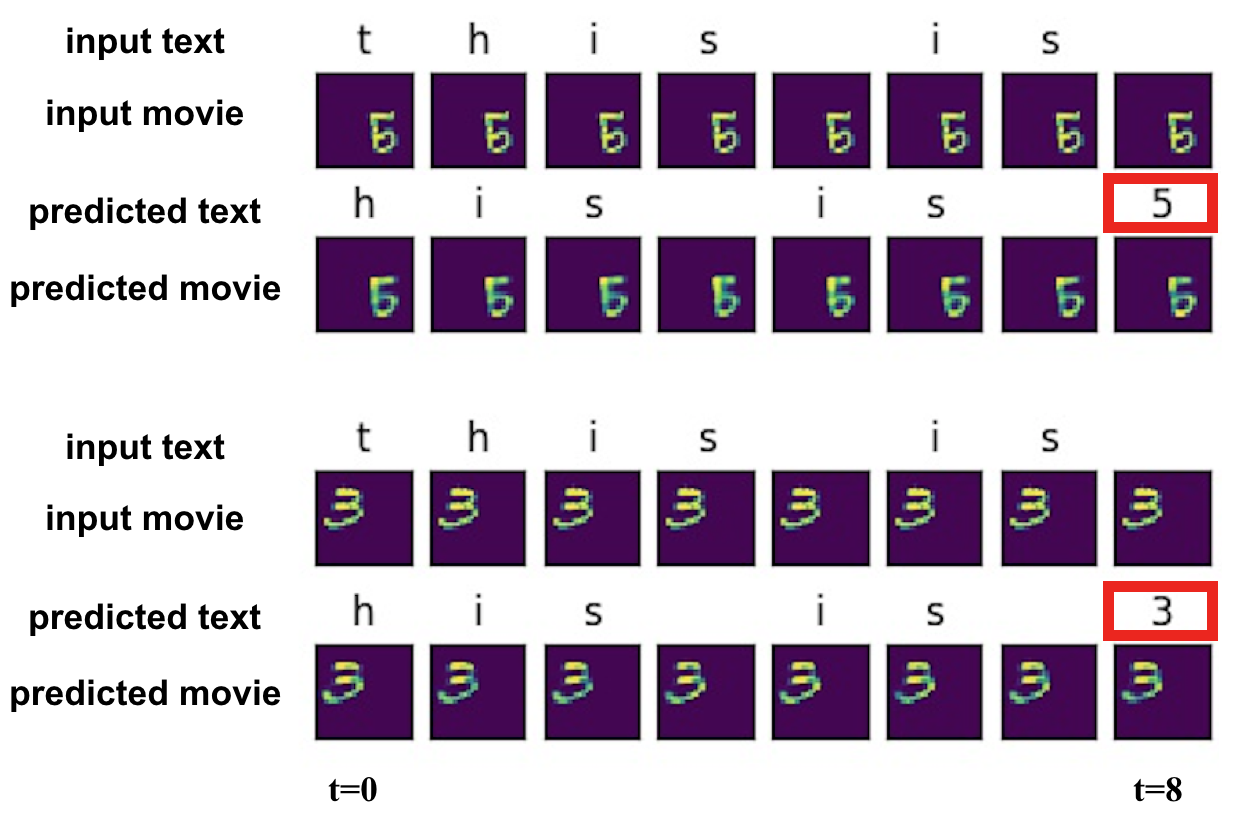}
  \caption[test1.]{LGI learns to classify digits with syntax ‘this is …’. LGI understood the meaning of the command text and managed to extract digit identity according to the morphology of digit instance. Note that the classification is performed by the proposed textizer rather than softmax.
}
\end{figure}

Based on the same network, LGI continued to learn syntax ‘this is …’. Just like a parent teaching child numbers by pointing to number instances, Figure 4 demonstrates that, after training of 50000 steps, LGI could classify figures in various morphology with correct identity (accuracy = 72.7\%). Note that, the classification process is not performed by softmax operation, but by directly textizing operation (i.e. rounding followed by a symbol mapping operation), which is more biologically plausible than the softmax operation.

\begin{figure}[!htbp]
  \centering\includegraphics[width=1\textwidth]{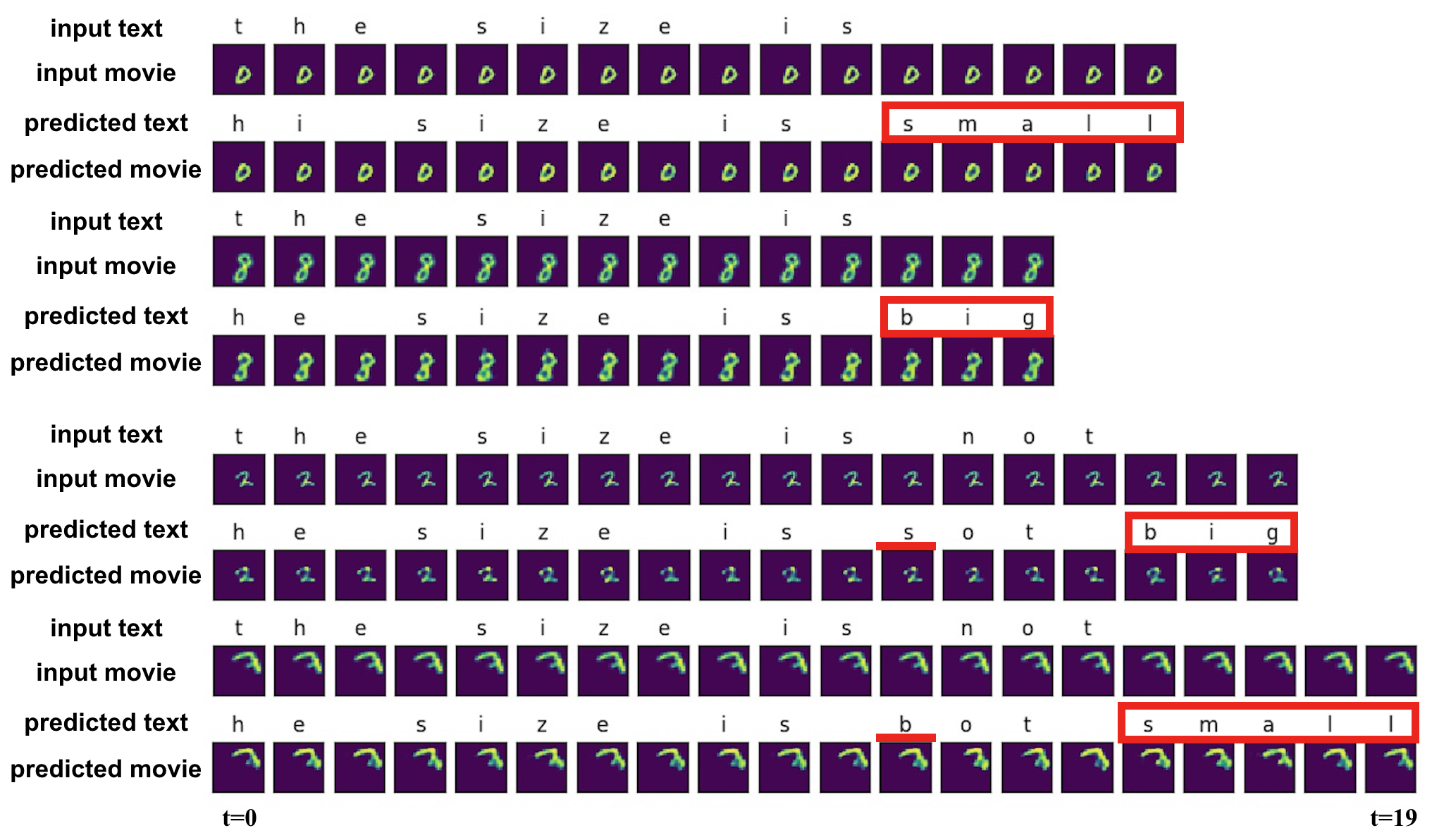}
  \caption[test1.]{LGI learns to judge the digit size with syntaxes ’the size is big/small’ and ’the size is not small/big’. LGI could understand the text command, offer correct judgment on digit size, and properly adjust the answer when encountered negative adverb ‘not’.
}
\end{figure}

After that, LGI learned the syntax ‘the size is big/small’, followed by ‘the size is not small/big’. Figure 5 illustrates that LGI could correctly categorize whether the digit size was small or big with proper text output. And we witness that, based on the syntax of ‘the size is big/small’ (train steps =1000), the negative adverb ‘not’ in the language text ‘the size is not small/big’ was much easier to be learned (train steps =200, with same hyper-parameters). This is quite similar to the cumulative learning process of the human being.

\begin{figure}[!htbp]
  \centering\includegraphics[width=1\textwidth]{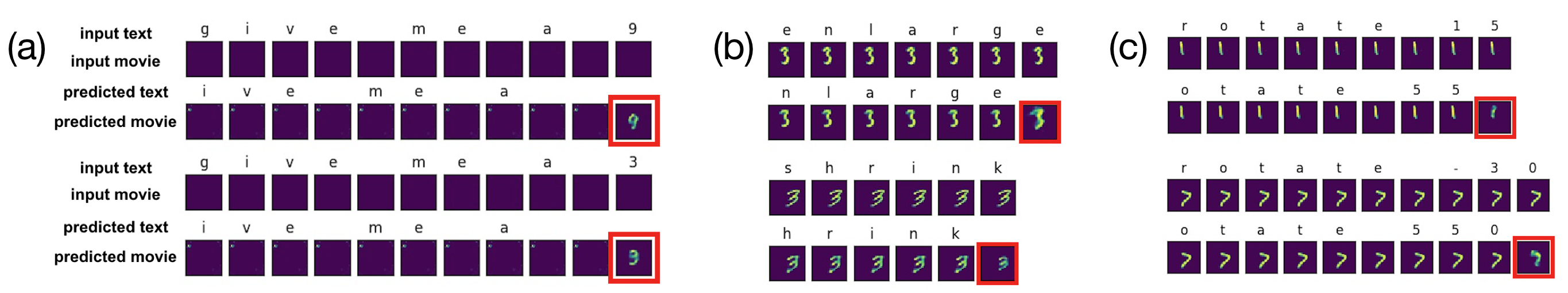}
  \caption[test1.]{LGI learns to generate a fictitious digit instance with syntax ‘give me a [number]’, and ‘mentally’ manipulate objects with syntaxes ‘enlarge’, ‘shrink’, and ‘rotate …’ etc. 

}
\end{figure}

And then, LGI rapidly learned three more syntaxes: ‘give me a …’, ‘enlarge/shrink’, and ‘rotate …’, whose results are shown in Figure 6. After training (5000 steps), LGI could generate a correct digit figure given the language command ‘give me a [number]’ (Figure 6.A). The generated digit instance is somewhat the ‘averaged’ version of all training examples of the same digit identity. In the future, the generative adversarial network (GAN) technique could be included to generate object instances with specific details. However, using more specific language, such as ‘give me a red Arial big 9’ to generate the characterized instance can better resemble the human thinking process than GAN. LGI can also learn to change the size and orientation of an imagined object. Figure 6.B-C illustrates the morphology of the final imagined instance could be kept unchanged after experiencing various manipulations. Some other syntaxes or tasks could be integrated into LGI in a similar way.

\begin{figure}[!htbp]
  \centering\includegraphics[width=1\textwidth]{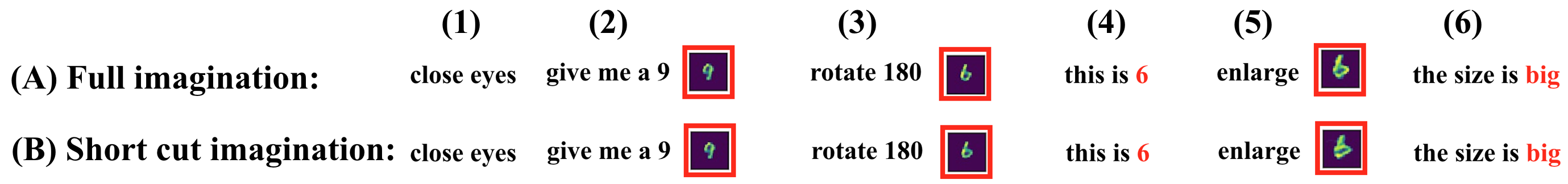}
  \caption[test1.]{The language guided thinking process. LGI generated an instance of digit ‘9’ without any input image. Then the instance was ‘mentally’ rotated 180 degree, based on which LGI found that the digit identity was changed to 6. After that, LGI enlarged the instance and identified its proper size.
}
\end{figure}

Finally, in Figure 7, we illustrate how LGI performed the human-like language-guided thinking process, with the above-learned syntaxes. (1) LGI first closed its eyes, namely, that no input images were fed into the vision subsystem (all the subsequent input images were generated through the imagination process). (2) LGI said to itself ‘give me a 9’, then the PFC produced the corresponding encoding vector \({\bf V_3’}\), and finally one digit ‘9’ instance was reconstructed via the imagination network. (3) LGI gave the command ‘rotate 180’, then the imagined digit ‘9’ was rotated upside down. (4) Following the language command ‘this is ’, LGI automatically predicted that the newly imaged object was the digit ‘6’. (5) LGI used ‘enlarge’ command to make the object bigger. (6) Finally, LGI predicted that the size was ‘big’ according to the imagined object morphology. This demonstrates that LGI can understand the verbs and nouns by properly manipulating the imagination, and can form the iterative thinking process via the interaction between vision and language subsystems through the PFC layer. The human thinking process normally would not form a concrete imagination through the full visual loop, but rather a vague and rapid imagination through the short-cut loop by feeding back \({\bf V_3’}\) to AIT directly. On the other hand, the full path of clear imagination may explain the dream mechanism. Figure 7.B shows the short cut imagination process, where LGI also regarded the rotated ‘9’ as digit 6, which suggests the AIT activation does not encode the digit identity, but the untangled features of input image or imagined image. Those high level cortices beyond visual cortex could be the place for identity representation.

\section{Discussion}
Language guided imagination is the nature of human thinking and intelligence. Normally, the real-time tasks or goals are conveyed by language, such as ‘to build a Lego car’. To achieve this goal, first, an agent (human being or machine) needs to know what’s car, and then imagine a vague car instance, based on which the agent can plan to later collect wheel, window and chassis blocks for construction. Imagining the vague car is the foundation for decomposing future tasks. We trained the LGI network with a human-like cumulative learning process, from learning the meaning of words, to understanding complicated syntaxes, and finally organizing the thinking process with language. We trained the LGI to associate object name with corresponding instances by ‘this is …’ syntax; and trained the LGI to produce a digit instance, when there comes the sentence ‘give me a [number]’. In contrast, traditional language models could only serve as a word dependency predictor rather than really understand the sentence.

Language is the most remarkable characteristics distinguishing mankind from animals. Theoretically, all kinds of information such as object properties, tasks and goals, commands and even emotions can be described and conveyed by language [21]. We trained with LGI eight different syntaxes (in other word, eight different tasks), and LGI demonstrates its understanding by correctly interacting with the vision system.  After learning ‘this is 9’, it is much easier to learn ‘give me a 9’; after learning the ‘size is big’, it is much easier to learn ‘the size is not small’. Maybe some digested words or syntaxes were represented by certain PFC units, which could be shared with the following sentence learning. 

Imagination is another key component of human thinking. For the game Go [22, 23], the network using a reinforcement learning strategy has to be trained with billions of games in order to acquire a feeling (Q value estimated for each potential action) to move the chess. As human beings, after knowing the rule conveyed by language, we can quickly start a game with proper moves using a try-in-imagination strategy without requiring even a single practice.  With imagination, people can change the answering contents (or even tell good-will lies) by considering or imagining the consequence of the next few output sentences. Machine equipped with the unique ability of imagination could easily select clever actions for multiple tasks without being trained heavily.

In the future, many more syntaxes and functionalities can be added to LGI in a similar way, such as math reasoning, intuitive physics prediction and navigation [24, 25, 26]. Insights of human audition processing could be leveraged to convert sound wave into language text as a direct input for LGI [27, 28]. And the mechanisms of human value systems in the striatum [29] may also endow LGI with motivation and emotion. The PFC cortex consists of many sub-regions interacted within the PFC and across the whole brain areas [3, 30], and the implementation of these features might finally enable LGI to possess real machine intelligence.

\section{Conclusion}

In this paper, we first introduced a PFC layer to involve representations from both language and vision subsystems to form a human-like thinking system (the LGI system). The LGI contains three subsystems: the vision, language, and PFC subsystem, which are trained separately. The development, recognition and learning mechanism is discussed in the cocurrent paper [10]. In the language subsystem, we use an LSTM layer to mimic the human IPS to extract the quantity information from language text and proposed a  biologically plausible textizer to produce text symbols output, instead of traditional softmax classifier. We propose to train the LGI with the NFP loss function, which endows the capacity to describe the image content in form of symbol text and manipulated images according to language commands. LGI shows its ability to learn eight different syntaxes or tasks in a cumulative learning way, and form the first machine thinking loop with the interaction between imagined pictures and language text. 
\section*{References}

\small
[1] Wei, M., He, Y., Zhang, Q. \ \& Si, L. (2019). Multi-Instance Learning for End-to-End Knowledge Base Question Answering.  {\it arXiv preprint} arXiv:1903.02652.

[2] Devlin, J., Chang, M. W., Lee, K. \ \& Toutanova, K. (2018). Bert: Pre-training of deep bidirectional transformers for language understanding. {\it arXiv preprint} arXiv:1810.04805.

[3] Miller, E. K. \ \& Cohen, J. D. (2001). An integrative theory of prefrontal cortex function. {\it Annual review of neuroscience}, {\bf24}(1), 167-202.

[4] Baddeley, A., Gathercole, S. \ \& Papagno, C. (1998). The phonological loop as a language learning device. {\it Psychological review}, {\bf 105}(1), 158.

[5] Finke, K., Bublak, P., Neugebauer, U. \ \& Zihl, J. (2005). Combined processing of what and where information within the visuospatial scratchpad. {\it European Journal of Cognitive Psychology}, {\bf 17}(1), 1-22.

[6] Simonyan, K. \ \& Zisserman, A. (2014). Very deep convolutional networks for large-scale image recognition. {\it arXiv preprint} arXiv:1409.1556.

[7] DiCarlo, J. J., Zoccolan, D. \ \& Rust, N. C. (2012). How does the brain solve visual object recognition?. {\it Neuron}, {\bf 73}(3), 415-434.

[8] Freiwald, W. A. \ \& Tsao, D. Y. (2010). Functional compartmentalization and viewpoint generalization within the macaque face-processing system. {\it Science}, {\bf 330}(6005), 845-851.

[9] Rosenblatt, F. (1958). The perceptron: a probabilistic model for information storage and organization in the brain. {\it Psychological review}, {\it 65}(6), 386.

[10] Anonymous A. (2019). The development, recognition, and learning mechanisms of animal-like neural network. {\it Advances in Neural Information Processing Systems, in submission}

[11] Rumelhart, D. E., Hinton, G. E. \ \& Williams, R. J. (1988). Learning representations by back-propagating errors. {\it Cognitive modeling}, {\bf 5}(3), 1.

[12] Yasuda, R., Sabatini, B. L. \ \& Svoboda, K. (2003). Plasticity of calcium channels in dendritic spines. {\it Nature neuroscience}, {\bf 6}(9), 948.

[13] Liu, L., Wong, T. P., Pozza, M. F., Lingenhoehl, K., Wang, Y., Sheng, M. \ \& Wang, Y. T. (2004). Role of NMDA receptor subtypes in governing the direction of hippocampal synaptic plasticity. {\it Science}, {\bf 304}(5673), 1021-1024.

[14] Pearson, J., Naselaris, T., Holmes, E. A. \ \& Kosslyn, S. M. (2015). Mental imagery: functional mechanisms and clinical applications. {\it Trends in cognitive sciences}, {\bf 19}(10), 590-602.

[15] Boureau, Y. L., Ponce, J. \ \& LeCun, Y. (2010). A theoretical analysis of feature pooling in visual recognition. {\it In Proceedings of the 27th international conference on machine learning} (ICML-10) (pp. 111-118).

[16] LeCun, Y., Bengio, Y. \ \& Hinton, G. (2015). Deep learning. {\it Nature}, {\bf 521}(7553), 436.

[17] Zhou, T., Brown, M., Snavely, N. \ \& Lowe, D. G. (2017). Unsupervised learning of depth and ego-motion from video. {\it In Proceedings of the IEEE Conference on Computer Vision and Pattern Recognition} (pp. 1851-1858).

[18] Ralph, M. A. L., Jefferies, E., Patterson, K. \ \& Rogers, T. T. (2017). The neural and computational bases of semantic cognition. {\it Nature Reviews Neuroscience}, {\bf 18}(1), 42.

[19] Petanjek, Z., Judaš, M., Kostović, I. \ \& Uylings, H. B. (2007). Lifespan alterations of basal dendritic trees of pyramidal neurons in the human prefrontal cortex: a layer-specific pattern. {\it Cerebral cortex}, {\bf 18}(4), 915-929.

[20] Goodfellow, I., Pouget-Abadie, J., Mirza, M., Xu, B., Warde-Farley, D., Ozair, S. \ \& Bengio, Y. (2014). Generative adversarial nets. {\it In Advances in neural information processing systems} (pp. 2672-2680).

[21] Wittgenstein, L. (2013). {\it Tractatus logico-philosophicus}.

[22] Silver, D., Huang, A., Maddison, C. J., Guez, A., Sifre, L., Van Den Driessche, G., \ \& Dieleman, S. (2016). Mastering the game of Go with deep neural networks and tree search. {\it Nature}, {\bf 529}(7587), 484.

[23] Silver, D., Schrittwieser, J., Simonyan, K., Antonoglou, I., Huang, A., Guez, A. \ \& Chen, Y. (2017). Mastering the game of go without human knowledge. {\it Nature}, {\bf 550}(7676), 354.

[24] Saxton, D., Grefenstette, E., Hill, F. \ \& Kohli, P. (2019). Analysing Mathematical Reasoning Abilities of Neural Models. {\it arXiv preprint} arXiv:1904.01557.

[25] Battaglia, P., Pascanu, R., Lai, M. \ \& Rezende, D. J. (2016). Interaction networks for learning about objects, relations and physics. {\it In Advances in neural information processing systems} (pp. 4502-4510).

[26] Banino, A., Barry, C., Uria, B., Blundell, C., Lillicrap, T., Mirowski, P. \ \& Wayne, G. (2018). Vector-based navigation using grid-like representations in artificial agents. {\it Nature}, {\bf 557}(7705), 429.

[27] Jasmin, K., Lima, C. F. \ \& Scott, S. K. (2019). Understanding rostral–caudal auditory cortex contributions to auditory perception. {\it Nature Reviews Neuroscience}, in press.

[28] Afouras, T., Chung, J. S. \ \& Zisserman, A. (2018). The conversation: Deep audio-visual speech enhancement. {\it arXiv preprint} arXiv:1804.04121.
 
[29] Husain, M. \ \& Roiser, J. (2018). Neuroscience of apathy and anhedonia: a transdiagnostic approach. {\it Nature Reviews Neuroscience}, {\bf 19}, 470-484.

[30] Barbas, H. (2015). General cortical and special prefrontal connections: principles from structure to function. {\it Annual review of neuroscience}, {\bf 38}, 269-289.

\end{document}